\documentclass[runningheads]{llncs}
\pdfoutput=1
\usepackage{graphicx}
\usepackage{amsmath,amssymb} 
\usepackage{color}
\usepackage[width=122mm,left=12mm,paperwidth=146mm,height=193mm,top=12mm,paperheight=217mm]{geometry}

\usepackage{url}
\usepackage[table]{xcolor}
\usepackage{bbm}
\usepackage{booktabs}
\usepackage[T1]{fontenc}
\usepackage{fix-cm}
\usepackage{array}
\usepackage{epsfig}
\usepackage{dsfont}
\usepackage{multirow}
\usepackage{hyperref}

\usepackage{times}
\usepackage{helvet}
\usepackage{courier}
\usepackage{graphicx}
\usepackage{bm}
\usepackage{color}
\usepackage{epstopdf}
\usepackage{caption}
\usepackage{subcaption}
\usepackage{enumitem}
\usepackage{calc}
\usepackage{multirow}
\usepackage{xspace}
\usepackage{booktabs}
\usepackage{mathrsfs}
\usepackage{array}
\usepackage{gensymb}

\newcommand{\figref}[1]{Fig\onedot~\ref{#1}}
\newcommand{\equref}[1]{Eq\onedot~\eqref{#1}}
\newcommand{\secref}[1]{Sec\onedot~\ref{#1}}
\newcommand{\tabref}[1]{Tab\onedot~\ref{#1}}

\newcommand{\ve}[1]{{\mathbf #1}} 

\newcommand{\spa}[1]{{\mathbb #1}}
\newcommand{\by}[2]{\ensuremath{#1 \! \times \! #2}}
\makeatletter
\newcommand{\printfnsymbol}[1]{%
  \textsuperscript{\@fnsymbol{#1}}%
}
\makeatother

\DeclareRobustCommand\onedot{\futurelet\@let@token\@onedot}
\def\onedot{\ifx\@let@token.\else.\null\fi\xspace}
\def\eg{\emph{e.g.}}

\def\ie{\emph{i.e.}}

\def\wrt{w.r.t\onedot} 

\def\etal{\emph{et al.}}

\begin{document}
\pagestyle{headings}
\mainmatter

\title{Depth Estimation via Affinity Learned with Convolutional Spatial Propagation Network} 

\titlerunning{CSPN}

\authorrunning{X. Cheng, P. Wang and R. Yang}

\author{Xinjing Cheng\thanks{equal contribution}, Peng Wang\printfnsymbol{1} and Ruigang Yang}


\institute{Baidu Research, Baidu Inc.\\
	\email{ \{chengxinjing,wangpeng54,yangruigang\}@baidu.com}
}

\maketitle

\begin{abstract}
Depth estimation from a single image is a fundamental problem in computer vision. In this paper, we propose a simple yet effective convolutional spatial propagation network (CSPN) to learn the affinity matrix for depth prediction. 
Specifically, we adopt an efficient linear propagation model, where the propagation is performed with a manner of recurrent convolutional operation, and the affinity among neighboring pixels is learned through a deep convolutional neural network (CNN). 
We apply the designed CSPN to two depth estimation tasks given a single image:  (1) Refine the depth output from existing state-of-the-art (SOTA)  methods;  (2) Convert sparse depth samples to a dense depth map by embedding the depth samples within the propagation procedure. The second task is inspired by the availability of LiDAR that provides sparse but accurate depth measurements. We experimented the proposed CSPN over the popular NYU v2~\cite{silberman2012indoor} and KITTI~\cite{geiger2012we} datasets, where we show that our proposed approach improves  not only quality (e.g., 30\% more reduction in depth error), but also speed (e.g., 2 to 5$\times$ faster) of depth maps than previous SOTA methods. 
\end{abstract}

\section{Introduction}
\label{sec:intro}
\begin{figure}[!htpb]
\centering
\includegraphics[width=1.02\textwidth]{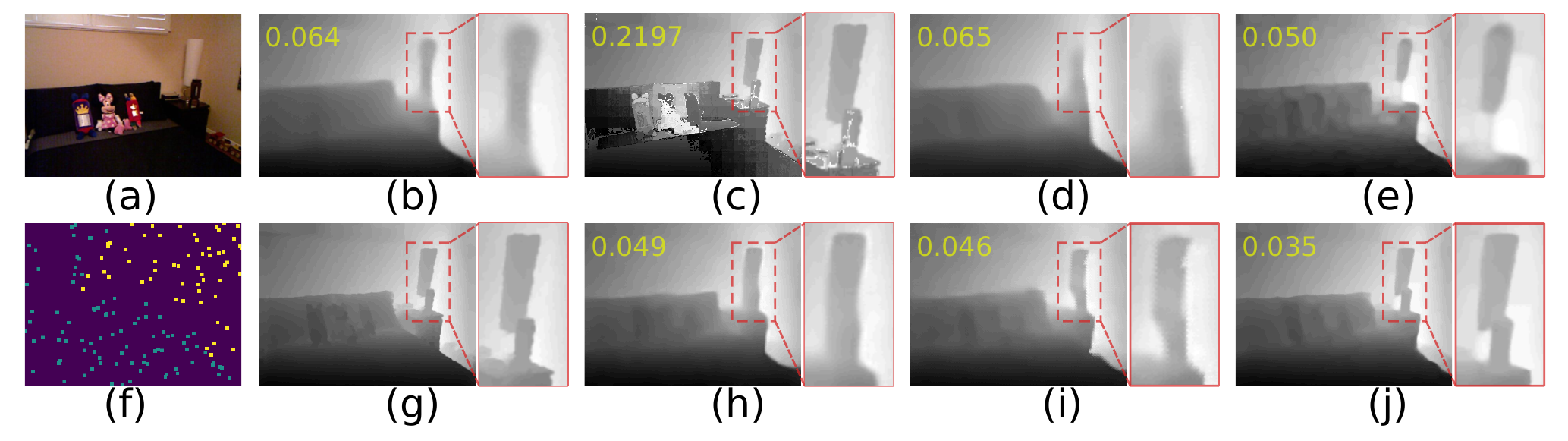}
\caption{(a) Input image; (b) Depth from~\cite{Ma2017SparseToDense}; (c) Depth after bilateral filtering; (d) Refined depth by SPN \cite{liu2017learning}; (e) Refined depth by CSPN; (f) Sparse depth samples (500); (g) Ground Truth; (h) Depth from our network; (i) Refined depth by SPN with depth sample; (j) Refined depth by CSPN with depth sample. The corresponding root mean square error (RMSE) is put at the left-top of each predicted depth map.}
\label{fig:example}
\end{figure}

Depth estimation from a single image, \ie, predicting per-pixel distance to the camera,  has many applications from augmented realities (AR), autonomous driving,  to robotics.  
Given a single image, recent efforts to estimate per-pixel depths have yielded high-quality outputs by taking advantage of deep fully convolutional neural networks~\cite{eigen2015predicting,laina2016deeper} and large amount of training data from indoor~\cite{silberman2012indoor,xiao2013sun3d,Matterport3D} and outdoor~\cite{geiger2012we,wang2016torontocity,huang2018apolloscape}. 
The improvement lies mostly in more accurate estimation of global scene layout and scales with advanced networks, such as VGG~\cite{simonyan2014very} and ResNet~\cite{HeZRS15}, and better local structure recovery through deconvolution operation~\cite{long2015fully}, skip-connections~\cite{ronneberger2015u} or up-projection~\cite{laina2016deeper}. 
Nevertheless, upon closer inspection of the output from a contemporary approach~\cite{Ma2017SparseToDense} (\figref{fig:example}(b)), the predicted depths is still blurry and do not align well with the given image structure such as object silhouette.

Most recently, Liu \etal~\cite{liu2017learning} propose to directly learn the image-dependent affinity through a deep CNN with spatial propagation networks (SPN), yielding better results comparing to the manually designed affinity on image segmentation. 
However, its propagation is performed in a scan-line or scan-column fashion, which is serial in nature. For instance, when propagating left-to-right, pixels at right-most column must wait the information from the left-most column to update its value. 
Intuitively, depth refinement commonly just needs a local context rather a global one. 

Here we propose convolutional spatial propagation networks (CSPN), where the depths at all pixels are updated simultaneously within a local convolutional context. The long range context is obtained through a recurrent operation. \figref{fig:example} shows an example, the depth estimated from CSPN (e) is more accurate than that from SPN (d) and Bilateral filtering (c). In our experiments, our parallel update scheme leads to significant performance improvement in both speed and quality over the serial ones such as SPN.

Practically, we show that the proposed strategy can also be easily extended to convert sparse depth samples to a dense depth map given corresponding image~\cite{LiaoHWKYL16,Ma2017SparseToDense}.  This task can be widely applied in robotics and autonomous cars, where depth perception is often acquired through LiDAR, which usually generates sparse but accurate depth measurement. By combining the sparse measurements with images, we could generate a full-frame dense depth map.
For this task, we consider three important requirements for an algorithm: 
(1) The dense depth map recovered should align with image structures;
(2) The depth value from the sparse samples should be preserved, since they are usually from a reliable sensor;
(3) The transition between sparse depth samples and their neighboring depths should be smooth and unnoticeable.  
In order to satisfy those requirements, we first add mirror connections based on the network from~\cite{Ma2017SparseToDense}, which generates better depths as shown in \figref{fig:example}(h). Then, we tried to embed the propagation into SPN in order to keep the depth value at sparse points. As shown in \figref{fig:example}(i), it generates better details and lower error than SPN without depth samples (\figref{fig:example}(d)). Finally, changing SPN to our CSPN yields the best result (\figref{fig:example}(j)). 
As can be seen, our recovered depth map with just 500 depth samples produces much more accurately estimated scene layouts and scales. 
We experiment our approach over two popular benchmarks for depth estimation, \ie NYU v2~\cite{silberman2012indoor} and KITTI~\cite{geiger2012we}, with standard evaluation criteria. In both datasets, our approach is significantly better (relative $30\%$ improvement in most key measurements) than previous deep learning based state-of-the-art (SOTA) algorithms~\cite{LiaoHWKYL16,Ma2017SparseToDense}. More importantly, it is very efficient yielding 2-5$\times$ acceleration comparing with SPN. In summary, this paper has the following contributions: 
\begin{enumerate}
    \item We propose convolutional spatial propagation networks (CSPN) which is more efficient and accurate for depth estimation than the previous SOTA propagation strategy~\cite{liu2017learning}, without sacrificing the theoretical guarantee.
    
    \item  We extend CSPN to the task of converting sparse depth samples to dense depth map by using the provided sparse depths into the propagation process. It guarantees that the sparse input depth values are preserved in the final depth map.  
    It runs in real-time, which is well suited for robotics and autonomous driving applications, where sparse depth measurement from LiDAR can be fused with image data.
    
\end{enumerate}

\section{Related Work}
\label{sec:related}
Depth estimating and enhancement/refinement have long been center problems for computer vision and robotics. Here we summarize those works in several aspects without enumerating them all due to space limitation.

\noindent\textbf{Single view depth estimation via CNN and CRF.}
Deep neural networks (DCN) developed in recent years
provide strong feature representation for 
per-pixel depth estimation from a single image. Numerous algorithms are developed through supervised methods~\cite{wang2015designing,eigen2015predicting,laina2016deeper,li2017two}, semi-supervised methods~\cite{kuznietsov2017semi} or unsupervised methods~\cite{godard2016unsupervised,zhou2017unsupervised,yang2018aaai,yang2018lego}.
and add in skip and mirror connections.
Others tried to improve the estimated details further by appending a conditional random field (CRF)~\cite{DBLP:conf/cvpr/WangSLCPY15,Liu_2015_CVPR,li2015depth} and joint training~\cite{crfasrnn_iccv2015,peng2016depth}. 
However, the affinity for measuring the coherence of neighboring pixels is manually designed.

\noindent\textbf{Depth Enhancement.}
Traditionally, depth output can be also efficiently enhanced with explicitly designed affinity through image filtering~\cite{barron2016fast,matsuo2015depth}, or data-driven ones through total variation (TV)~\cite{ferstl2013image,ferstl2015variational} and learning to diffuse~\cite{liu2016learning} by incorporating more priors into diffusion partial differential equations (PDEs).
However, due to the lack of an effective learning strategy, they are limited for large-scale complex visual enhancement.

Recently, deep learning based enhancement yields impressive results on super resolution of both images~\cite{dong2014learning,yang2014color} and depths~\cite{song2016deep,hui2016depth,kwon2015data,riegler2016atgv}. The network takes low resolution inputs and output the high-resolution results, and is trained end-to-end where the mapping between input and output is implicitly learned.
However, these methods are only trained and experimented with perfect correspondent ground-truth low-resolution and high-resolution depth maps and often a black-box model. In our scenario, both the input and ground truth depth are non-perfect, \eg depths from a low cost LiDAR or a network, thus an explicit diffusion process to guide the enhancement such as SPN is necessary.

\noindent\textbf{Learning affinity for spatial diffusion.}
Learning affinity matrix with deep CNN for diffusion or spatial propagation receives high interests in recent years due to its theoretical supports and guarantees~\cite{weickert1998anisotropic}.
Maire \etal~\cite{maire2016affinity} trained a deep CNN to directly predict the entities of an affinity matrix, which demonstrated good performance on image segmentation. However, the affinity is followed by an independent non-differentiable solver of spectral embedding, it can not be supervised end-to-end for the prediction task. Bertasius \etal~\cite{bertasius2016convolutional} introduced a random walk network that optimizes the objectives of pixel-wise affinity for semantic segmentation. Nevertheless, their affinity matrix needs additional supervision from ground-truth sparse pixel pairs, which limits the potential connections between pixels. Chen \etal~\cite{chen2016semantic} try to explicit model an edge map for domain transform to improve the output of neural network. 

The most related work with our approach is SPN~\cite{liu2017learning}, where the learning of a large affinity matrix for diffusion is converted to learning a local linear spatial propagation, yielding a simple while effective approach for output enhancement. However, as mentioned in~\secref{sec:intro}, depth enhancement commonly needs local context, it might not be necessary to update a pixel by scanning the whole image. As shown in our experiments, our proposed CSPN is more efficient and provides much better results.

\noindent\textbf{Depth estimation with given sparse samples.}
The task of sparse depth to dense depth estimation was introduced in robotics due to its wide application for enhancing 3D perception~\cite{LiaoHWKYL16}. Different from depth enhancement, the provided depths are usually from low-cost LiDAR or one line laser sensors, yielding a map with valid depth in only few hundreds of pixels, as illustrated in \figref{fig:example}(f). 
Most recently, Ma \etal~\cite{Ma2017SparseToDense} propose to treat sparse depth map as additional input to a ResNet~\cite{laina2016deeper} based depth predictor, producing superior results than the depth output from CNN with solely image input. However, the output results are still blurry, and does not satisfy our requirements of depth as discussed in \secref{sec:intro}. In our case, we directly embed the sampled depth in the diffusion process, where all the requirements are held and guaranteed.

Some other works directly convert sparse 3D points to dense ones without image input~\cite{Zimmermann2017Learning,Ladicky_2017_ICCV,uhrig2017sparsity}, whereas the density of sparse points must be high enough to reveal the scene structure, which is not available in our scenario.
\section{Our Approach}
We formulate the problem as an anisotropic diffusion process and the diffusion tensor is learned through a deep CNN directly from the given image, which guides the refinement of the output.

\begin{figure}[t]
\includegraphics[width=1.0\textwidth]{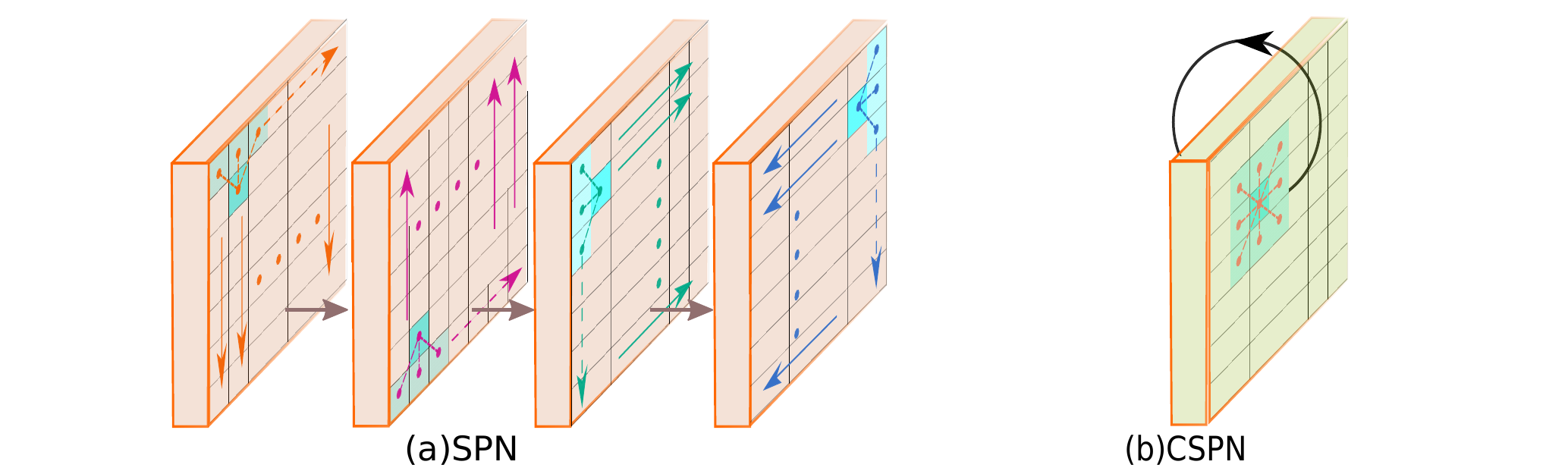}
\caption{Comparison between the propagation process in SPN~\cite{liu2017learning} and CPSN in this work.}
\label{fig:compare}
\end{figure}

\subsection{Convolutional Spatial Propagation Network}
Given a depth map $D_o \in \spa{R}^{m\times n}$ that is output from a network, and image $\ve{X} \in \spa{R}^{m\times n}$, our task is to update the depth map to a new depth map $D_n$ within $N$ iteration steps, which first reveals more details of the image, and second improves the per-pixel depth estimation results. 

\figref{fig:compare}(b) illustrates our updating operation. Formally, without loss of generality, we can embed the $D_o$ to some hidden space $\ve{H} \in \spa{R}^{m \times n \times c}$. The convolutional transformation functional with a kernel size of $k$ for each time step $t$ could be written as,
\begin{align}
    \ve{H}_{i, j, t + 1} &= \sum\nolimits_{a,b = -(k-1)/2}^{(k-1)/2} \kappa_{i,j}(a, b) \odot \ve{H}_{i-a, j-b, t} \nonumber \\
\mbox{where,~~~~}
    \kappa_{i,j}(a, b) &= \frac{\hat{\kappa}_{i,j}(a, b)}{\sum_{a,b, a, b \neq 0} |\hat{\kappa}_{i,j}(a, b)|}, \nonumber\\
    \kappa_{i,j}(0, 0) &= 1 - \sum\nolimits_{a,b, a, b \neq 0}\kappa_{i,j}(a, b)
\label{eqn:cspn}
\end{align}
where the transformation kernel $\hat{\kappa}_{i,j} \in \spa{R}^{k\times k \times c}$ is the output from an affinity network, which is spatially dependent on the input image. The kernel size $k$ is usually set as an odd number so that the computational context surrounding pixel $(i, j)$ is symmetric.
$\odot$ is element-wise product. Following~\cite{liu2017learning}, we normalize kernel weights between range of $(-1, 1)$ so that the model can be stabilized and converged by satisfying the condition $\sum_{a,b, a,b \neq 0} |\kappa_{i,j}(a, b)| \leq 1$. Finally, we perform this iteration $N$ steps to reach a stationary distribution.

\noindent\textbf{Correspondence to diffusion process with a partial differential equation (PDE).} \\
Similar with~\cite{liu2017learning}, here we show that our CSPN holds all the desired properties of SPN.
Formally, we can rewrite the propagation in \equref{eqn:cspn} as a process of diffusion evolution by first doing column-first vectorization of feature map $\ve{H}$ to $\ve{H}_v \in \spa{R}^{\by{mn}{c}}$.
\begin{align}
     \ve{H}_v^{t+1} = 
     \begin{bmatrix}
    1-\lambda_{0, 0}  & \kappa_{0,0}(1,0) & \cdots & 0 \\
    \kappa_{1,0}(-1,0)   & 1-\lambda_{1, 0} & \cdots & 0 \\
    \vdots & \vdots & \ddots & \vdots \\
    \vdots & \cdots & \cdots & 1-\lambda_{m,n} \\
\end{bmatrix} = \ve{G}\ve{H}_v^{t}
\label{eqn:vector}
\end{align}
where $\lambda_{i, j} = \sum_{a,b}\kappa_{i,j}(a,b)$ and $\ve{G}$ is a $\by{mn}{mn}$ transformation matrix. The diffusion process expressed with a partial differential equation (PDE) is derived as follows, 
\begin{align}
     \ve{H}_v^{t+1} &= \ve{G}\ve{H}_v^{t} = (\ve{I} - \ve{D} + \ve{A})\ve{H}_v^{t} \nonumber\\
     \ve{H}_v^{t+1} - \ve{H}_v^{t} &= - (\ve{D} - \ve{A}) \ve{H}_v^{t} \nonumber\\
     \partial_t \ve{H}_v^{t+1} &= -\ve{L}\ve{H}_v^{t}
\label{eqn:proof}
\end{align}
where $\ve{L}$ is the Laplacian matrix, $\ve{D}$ is the diagonal matrix containing all the $\lambda_{i, j}$, and $\ve{A}$ is the affinity matrix which is the off diagonal part of $\ve{G}$.

In our formulation, different from~\cite{liu2017learning} which scans the whole image in four directions~(\figref{fig:compare}(a)) sequentially, CSPN propagates a local area towards all directions at each step~(\figref{fig:compare}(b)) simultaneously, \ie with~\by{k}{k} local context, while larger context is observed when recurrent processing is performed, and the context acquiring rate is in an order of $O(kN)$.

In practical, we choose to use convolutional operation due to that it can be efficiently implemented through image vectorization, yielding real-time performance in depth refinement tasks.

Principally, CSPN could also be derived from loopy belief propagation with sum-product algorithm~\cite{kschischang2001factor}. However, since our approach adopts linear propagation, which is efficient while just a special case of pairwise potential with L2 reconstruction loss in graphical models. Therefore, to make it more accurate, we call our strategy convolutional spatial propagation in the field of diffusion process.

\begin{figure}[t]
\centering
\includegraphics[width=0.9\textwidth]{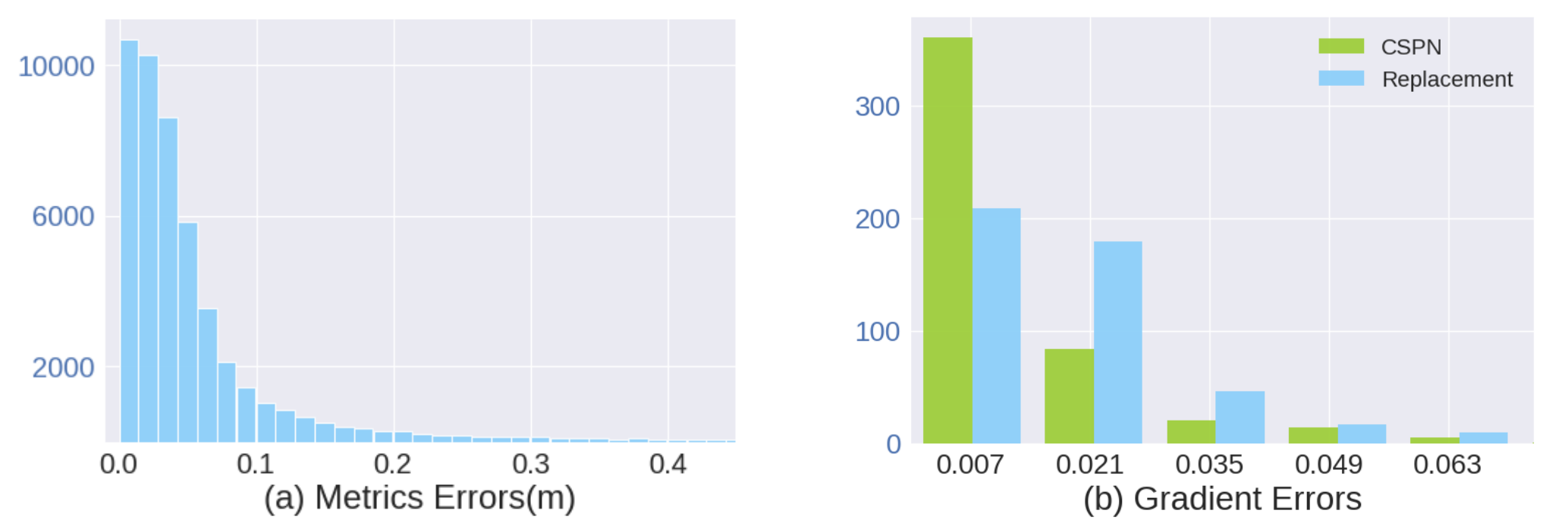}
\caption {(a) Histogram of RMSE with depth maps from~\cite{Ma2017SparseToDense} at given sparse depth points.  (b) Comparison of gradient error between depth maps with sparse depth replacement (blue bars) and with ours CSPN (green bars), where ours is much smaller. Check~\figref{fig:gradient} for an example. Vertical axis shows the count of pixels.}
\label{fig:hist}
\end{figure}

\subsection{Spatial Propagation with Sparse Depth Samples}
\begin{figure}[t]
\centering
\includegraphics[width=0.95\textwidth]{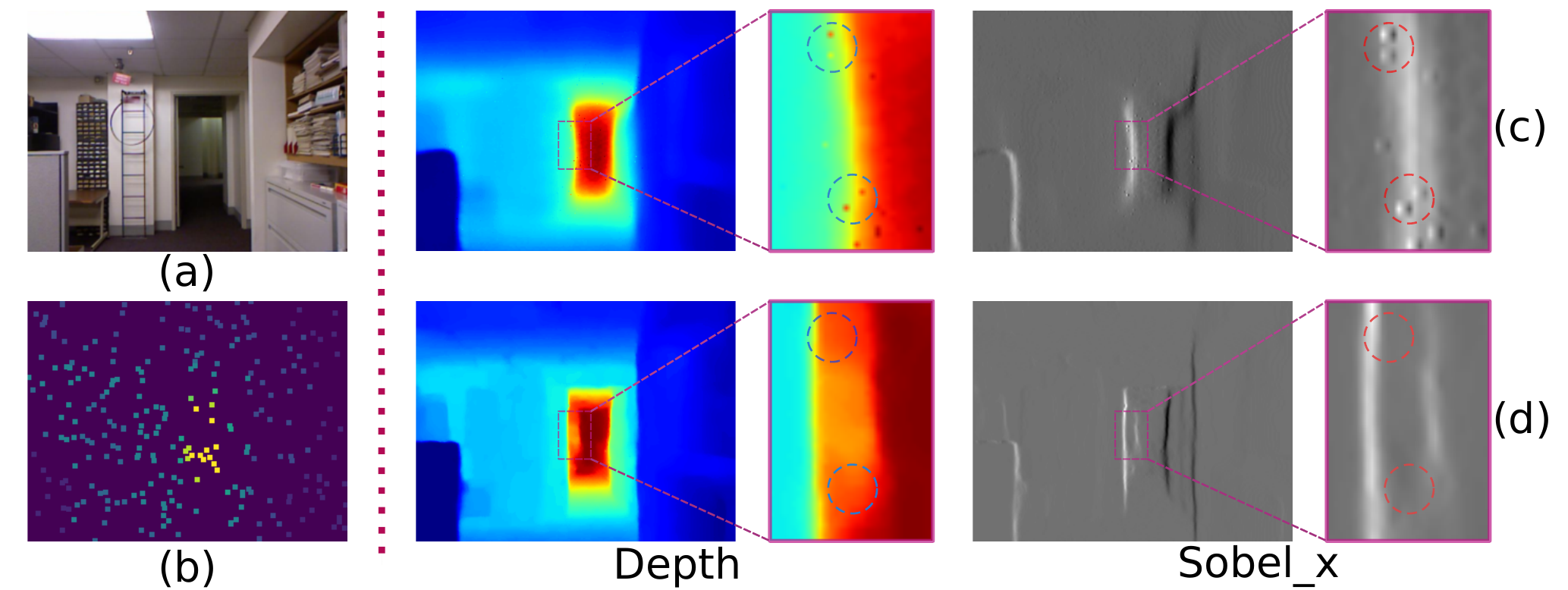}
\caption{Comparison of depth map~\cite{Ma2017SparseToDense} with sparse depth replacement and with our CSPN \wrt smoothness of depth gradient at sparse depth points. (a) Input image. (b) Sparse depth points. (c) Depth map with sparse depth replacement. (d) Depth map with our CSPN with sparse depth points. We highlight the differences in the red box.}
\label{fig:gradient}
\end{figure}

In this application, we have an additional sparse depth map $D_s$ (\figref{fig:gradient}(b)) to help estimate a depth depth map from a RGB image. Specifically, a sparse set of pixels are set with real depth values from some depth sensors, which can be used to guide our propagation process. 

Similarly, we also embed the sparse depth map $D_s = \{d_{i,j}^s\}$ to a hidden representation $\ve{H}^s$,  and we can write the updating equation of $\ve{H}$ by simply adding a replacement step after performing \equref{eqn:cspn}, 
\begin{align}
    \ve{H}_{i, j, t+1} = (1 - m_{i, j}) \ve{H}_{i, j, t+1}  +  m_{i, j} \ve{H}_{i, j}^s 
\label{eqn:cspn_sp}
\end{align}
where $m_{i, j} = \spa{I}(d_{i, j}^s > 0)$ is an indicator for the availability of sparse depth at $(i, j)$. 

In this way, we guarantee that our refined depths have the exact same value at those valid pixels in sparse depth map. Additionally, we propagate the information from those sparse depth to its surrounding pixels such that the smoothness between the sparse depths and their neighbors are maintained. 
Thirdly, thanks to the diffusion process, the final depth map is well aligned with image structures. 
This fully satisfies the desired three properties for this task which is discussed in our introduction (\ref{sec:intro}). 


In addition, this process is still following the diffusion process with PDE, where the transformation matrix can be built by simply replacing the rows satisfying $m_{i, j} = 1$ in $\ve{G}$ (\equref{eqn:vector}), which are corresponding to sparse depth samples, by $\ve{e}_{i + j*m}^T$. Here $\ve{e}_{i + j*m}$ is an unit vector with the value at $i + j*m$ as 1.
Therefore, the summation of each row is still $1$, and obviously the stabilization still holds in this case.

Our strategy has several advantages over the previous state-of-the-art sparse-to-dense methods~\cite{Ma2017SparseToDense,LiaoHWKYL16}.
In \figref{fig:hist}(a), we plot a histogram of depth displacement from ground truth at given sparse depth pixels from the output of Ma \etal~\cite{Ma2017SparseToDense}. It shows the accuracy of sparse depth points cannot preserved, and some pixels could have very large displacement (0.2m), indicating that directly training a CNN for depth prediction does not preserve the value of real sparse depths provided. To acquire such property, 
one may simply replace the depths from the outputs with provided sparse depths at those pixels, however, it yields non-smooth depth gradient \wrt surrounding pixels. 
In~\figref{fig:gradient}(c), we plot such an example, at right of the figure, we compute Sobel gradient~\cite{sobel2014history} of the depth map along x direction, where we can clearly see that the gradients surrounding pixels with replaced depth values are non-smooth.
We statistically verify this in \figref{fig:hist}(b) using 500 sparse samples, the blue bars are the histogram of gradient error  at sparse pixels by comparing the gradient of the depth map with sparse depth replacement and of ground truth depth map. We can see the difference is significant, 2/3 of the sparse pixels has large gradient error.
Our method, on the other hand, as shown with the green bars in \figref{fig:hist}(b), the average gradient error is much smaller, and most pixels have zero error. In\figref{fig:gradient}(d), we show the depth gradients surrounding sparse pixels are smooth and close to ground truth, demonstrating the effectiveness of our propagation scheme. 


\subsection{Complexity Analysis}
\label{subsec:time}

As formulated in~\equref{eqn:cspn}, our CSPN takes the operation of convolution, where the complexity of using CUDA with GPU for one step CSPN is $O(\log_2(k^2))$, where $k$ is the kernel size. This is because CUDA uses parallel sum reduction, which has logarithmic complexity. In addition,  convolution operation can be performed parallel for all pixels and channels, which has a constant complexity of $O(1)$. Therefore, performing $N$-step propagation, the overall complexity for CSPN is $O(\log_2(k^2)N)$, which is irrelevant to image size $(m, n)$.

SPN~\cite{liu2017learning} adopts scanning row/column-wise propagation in four directions. Using $k$-way connection and running in parallel, the complexity for one step is $O(\log_2(k))$. The propagation needs to scan full image from one side to another, thus the complexity for SPN is $O(\log_2(k)(m + n))$. Though this is already more efficient than the densely connected CRF proposed by~\cite{philipp2012dense}, whose implementation complexity with permutohedral lattice is $O(mnN)$, ours $O(\log_2(k^2)N)$ is more efficient since the number of iterations $N$ is always much smaller than the size of image $m, n$. We show in our experiments (\secref{sec:exp}), with $k=3$ and $N=12$, CSPN already outperforms SPN with a large margin (relative $30\%$), demonstrating both efficiency and effectiveness of the proposed approach.

\subsection{End-to-End Architecture}
\label{subsec:unet}
\begin{figure}[t]
\centering
\includegraphics[width=0.95\textwidth,height=0.45\textwidth]{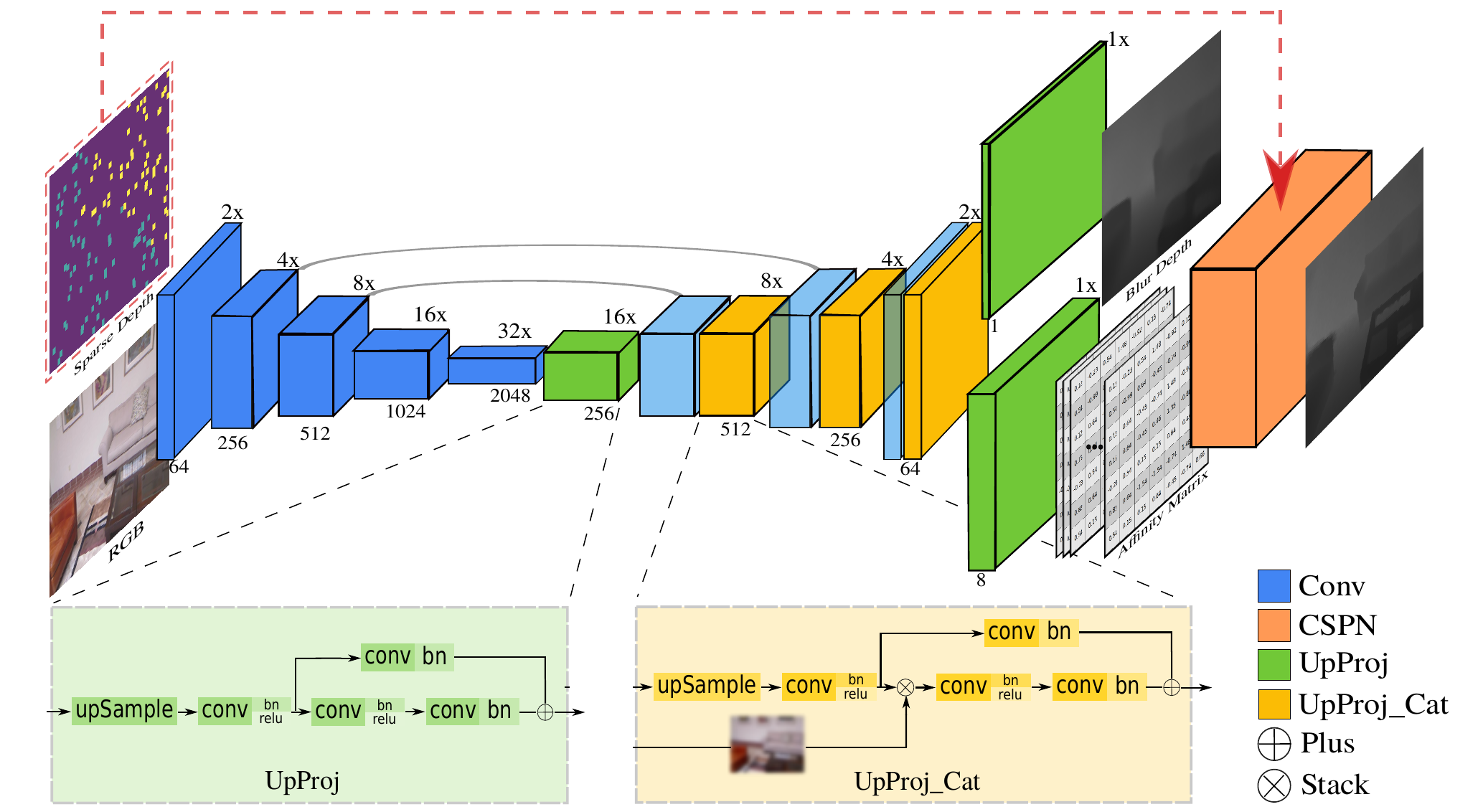}
\caption{Architecture of our networks with mirror connections for  depth estimation via transformation kernel prediction with CSPN (best view in color). Sparse depth is an optional input, which can be embedded into the CSPN to guide the depth refinement.}
\label{fig:arch}
\end{figure}

We now explain our end-to-end network architecture to predict both the transformation kernel and the depth value, which are the inputs to CSPN for depth refinement.
 As shown in \figref{fig:arch}, our network has some similarity with that from Ma \etal~\cite{Ma2017SparseToDense}, with the final CSPN layer that outputs a dense depth map.  
 
For predicting the transformation kernel $\kappa$ in \equref{eqn:cspn}, 
rather than building a new deep network for learning affinity same as Liu \etal~\cite{liu2017learning}, we branch an additional output from the given network, which shares the same feature extractor with the depth network. This helps us to save memory and time cost for joint learning of both depth estimation and transformation kernels prediction. 

Learning of affinity is dependent on fine grained spatial details of the input image. However, spatial information is weaken or lost with the down sampling operation during the forward process of the ResNet in~\cite{laina2016deeper}. Thus, we add mirror connections similar with the U-shape network~\cite{ronneberger2015u} by directed concatenating the feature from encoder to up-projection layers as illustrated by ``UpProj$\_$Cat'' layer in~\figref{fig:arch}. Notice that it is important to carefully select the end-point of mirror connections. Through experimenting three possible positions to append the connection, \ie after \textit{conv}, after \textit{bn} and after \textit{relu} as shown by the ``UpProj'' layer in~\figref{fig:arch} , we found the last position provides the best results by validating with the NYU v2 dataset (\secref{subsec:ablation}). 
In doing so, we found not only the depth output from the network is better recovered, and the results after the CSPN is additionally refined, which we will show the experiment section~(\secref{sec:exp}).
Finally we adopt the same training loss as~\cite{Ma2017SparseToDense}, yielding an end-to-end learning system.

\section{Experiments}
\label{sec:exp}
In this section, we describe our implementation details, the datasets and evaluation metrics used in our experiments. Then present comprehensive evaluation of CSPN on both depth refinement and sparse to dense tasks. 

\noindent\textbf{Implementation details.}
The weights of ResNet in the encoding layers for depth estimation (\secref{subsec:unet}) are initialized with models pretrained on the
ImageNet dataset~\cite{deng2009imagenet}.  
Our models are trained with SGD optimizer, and we use a small batch size of 24 and train for 40 epochs for all the experiments, and the model performed best on the validation set is used for testing.
The learning rate starts at 0.01, and is reduced to 20$\%$ every 10 epochs. A small weight decay of $10^{-4}$ is applied for regularization.
We implement our networks based on PyTorch~\footnote{http://pytorch.org/} platform, and use its element-wise product and convolution operation for our one step CSPN implementation.

For depth, we show that propagation with hidden representation $\ve{H}$ only achieves marginal improvement over doing propagation within the domain of depth $D$. Therefore, we perform all our experiments direct with $D$ rather than learning an additional embedding layer. For sparse depth samples, we adopt 500 sparse samples as that is used in~\cite{Ma2017SparseToDense}.

\subsection{Datasets and Metrics}
\label{subsec:data_metric}
All our experiments are evaluated on two datasets: NYU v2~\cite{silberman2012indoor} and KITTI ~\cite{geiger2012we}, using commonly used metrics.\\
\noindent\textbf{NYU v2.} The NYU-Depth-v2 dataset consists of RGB and depth images collected from 464 different indoor scenes. We use the official split of data, where 249 scenes are used for training and we sample 50K images out of the training set with the same manner as~\cite{Ma2017SparseToDense}. For testing, following the standard setting~\cite{eigen2015predicting,peng2016depth}, the small labeled test set with 654 images is used the final performance. The original image of size $\by{640}{480}$ are first downsampled to half and then center-cropped, producing a network input size of $\by{304}{228}$.\\
\noindent\textbf{KITTI odometry dataset.} It includes both camera and LiDAR measurements, and consists of 22 sequences. Half of the sequence is
used for training while the other half is for evaluation. Following~\cite{Ma2017SparseToDense}, we use all 46k images from the training sequences for training, and a random subset of 3200 images from the test sequences for evaluation. Specifically, we take the bottom part $\by{912}{228}$ due to no depth at the top area, and only evaluate the pixels with ground truth.\\
\noindent\textbf{Metrics.} We adopt the same metrics and use their implementation in \cite{Ma2017SparseToDense}. Given ground truth depth $D^* = \{d^*\}$ and predicted depth $D = \{d\}$, the metrics include:  (1) RMSE: $\sqrt{\frac{1}{|D|}\sum_{d \in D}||d^* - d||^2}$. (2) Abs Rel: $\frac{1}{|D|}\sum_{d \in D}|d^* - d|/d^*$. (3) $\delta_t$: $\%$ of $d \in D$, s.t. $max(\frac{d^*}{d}, \frac{d}{d^*})<t$, where $t \in \{1.25, 1.25^2, 1.25^3\}$. Nevertheless, for the third metric, we found that the depth accuracy is very high when sparse depth is provided, $t = 1.25$ is already a very loosen criteria where  almost $100\%$ of pixels are judged as correct, which can hardly distinguish different methods as shown in (\tabref{tbl:sota}). Thus we adopt more strict criteria for correctness by choosing $t \in \{1.02, 1.05, 1.10\}$.





\begin{figure}[t]
\centering
\includegraphics[width=1.0\textwidth]{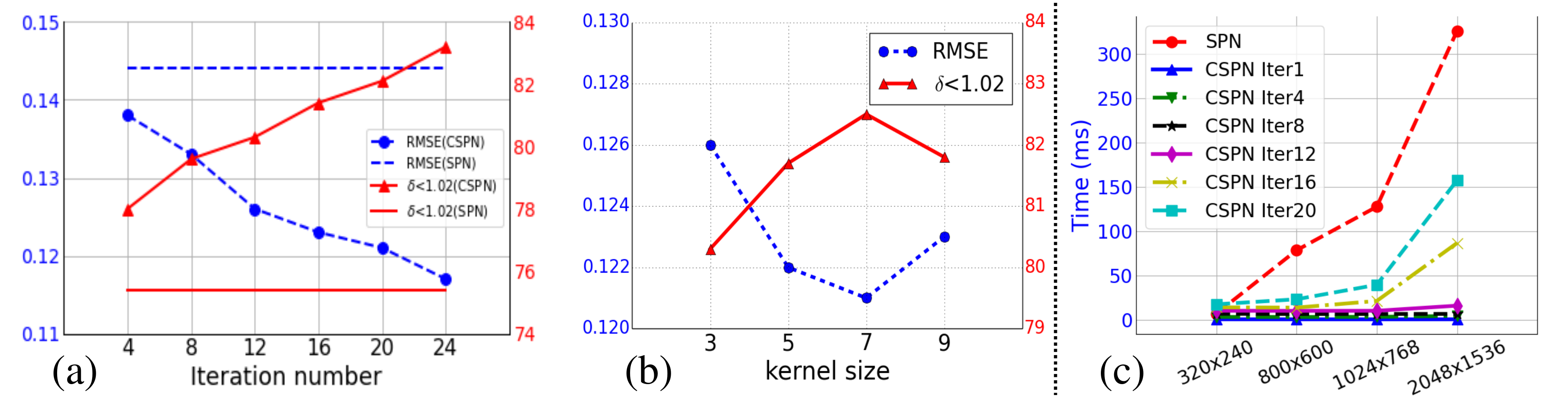}
\caption{Ablation study.(a) RMSE (left axis, lower the better) and $\delta < 1.02$ (right axis, higher the better) of CSPN \wrt number of iterations. Horizontal lines show the corresponding results from SPN~\cite{liu2017learning}. (b) RMSE and $\delta < 1.02$ of CSPN \wrt kernel size. (c) Testing times \wrt input image size.}
\label{fig:ab_study}
\end{figure}

\subsection{Parameter Tuning and Speed Study}
\label{subsec:ablation}
We first evaluate various hyper-parameters including kernel size $k$, number of iterations $N$ in \equref{eqn:cspn} using the NYU v2 dataset. Then we provide an empirical evaluation of the running speed with a Titan X GPU on a computer with 16 GB memory.

\noindent\textbf{Number of iterations.} We adopt a kernel size of $3$ to validate the effect of iteration number $N$ in CSPN. 
As shown in  \figref{fig:ab_study}(a), our CSPN has outperformed SPN~\cite{liu2017learning} (horizontal line) when iterated only four times. Also, we can get even better performance when more iterations are applied in the model during training. From our experiments, the accuracy is saturated when the number of iterations is increased to $24$. 

\noindent\textbf{Size of convolutional kernel.} As shown in  \figref{fig:ab_study}(b), larger convolutional kernel has similar effect with more iterations, due to larger context is considered for propagation at each time step. Here, we hold the iteration number to $N = 12$, and we can see the performance is better when $k$ is larger while saturated at size of $7$. 
We notice that the performance drop slightly when kernel size is set to $9$. This is because we use a fixed number of epoch, \ie 40, for all the experiments, while larger kernel size induces more affinity to learn in propagation, which needs more epoch of data to converge. Later, when we train with more epochs, the model reaches similar performance with kernel size of $7$.
Thus, we can see using kernel size of $7$ with $12$ iterations reaches similar performance of using kernel size of $3$ with $20$ iterations, which shows CSPN has the trade-off between kernel size and iterations. In practice, the two settings run with similar speed, while the latter costs much less memory. Therefore, we adopt kernel size as $3$ and number of iterations as $24$ in our comparisons.

\noindent\textbf{Concatenation end-point for mirror connection.} 
As discussed in \secref{subsec:unet}, based on the given metrics, we experimented three concatenation places, \ie after \textit{conv}, after \textit{bn} and after \textit{relu} by fine-tuning with weights initialized from encoder network trained without mirror-connections.
The corresponding RMSE are  $0.531$, $0.158$ and $0.137$ correspondingly. Therefore, we adopt the proposed concatenation end-point.

\noindent\textbf{Running speed}
In  \figref{fig:ab_study}(c), we show the running time comparison between the SPN and CSPN with kernel size as $3$. We use the author's PyTorch implementation online. As can be seen, we can get better performance within much less time. For example, four iterations of CSPN on one $1024\times768$ image only takes 3.689~$ms$, while SPN takes 127.902~$ms$. In addition, the time cost of SPN is linearly growing \wrt image size, while the time cost of CSPN is irrelevant to image size and much faster as analyzed in \secref{subsec:time}. In practice, however, when the number of iterations is large, \eg ``CSPN Iter 20'', we found the practical time cost of CSPN also grows \wrt image size. This is because of PyTorch-based implementation, which keeps all the variables for each iteration in memory during the testing phase. Memory paging cost becomes dominant with large images. In principle, we can eliminate such a memory bottleneck by customizing a new operation, which will be our future work. Nevertheless, without coding optimation, even at high iterations with large images, CSPN's speed is still twice as fast as SPN. 

\begin{table}[t]
	\centering
	\fontsize{7.5}{7.5}\selectfont
	\caption{Comparison results on NYU v2 dataset~\cite{silberman2012indoor} between different variants of CSPN and other state-of-the-art strategies. Here, ``Preserve SD'' is short for preserving the depth value at sparse depth samples.}
	\bgroup
	\def\arraystretch{1.3}
	\setlength{\tabcolsep}{4.5pt} 
	\begin{tabular}{lccccccccc}
		\hline
		\multicolumn{1}{l}{\multirow{2}{*}{Method}}  & \multirow{2}{*}{Preserve ``SD''} & \multicolumn{2}{c}{Lower the better}   & \multicolumn{6}{c}{Higher the better} \\ \cline{3-10}
		\multicolumn{1}{l}{} & & RMSE  & \multicolumn{1}{c|}{REL} & $\delta_{1.02}$ & $\delta_{1.05}$ & $\delta_{1.10}$ & $\delta_{1.25}$ & $\delta_{1.25^2}$ & $\delta_{1.25^3}$ \\ \hline
		~~Ma \etal \cite{Ma2017SparseToDense} &   & 0.230  & 0.044    & 52.3            & 82.3            & 92.6            & 97.1            & 99.4              & 99.8              \\ \hline
		+Bilateral~\cite{barron2016fast}&   & 0.479  & 0.084    & 29.9            & 58.0            & 77.3            & 92.4            & 97.6              & 98.9     \\\hline
		+SPN~\cite{liu2016learning}      &  & 0.172          & 0.031                    & 61.1            & 84.9            & 93.5            & 98.3            & 99.7              & 99.9              \\ \hline
	    +CSPN (Ours)     &  & 0.162          & 0.028                    & 64.6            & 87.7            & 94.9            & 98.6            & 99.7              & 99.9              \\\hline
	    +UNet (Ours)     &  & 0.137          & 0.020                    & 78.1       & 91.6 & 96.2 & 98.9 & 99.8 & 100.0              \\ \hline
        \hline
        +ASAP~\cite{igarashi2005rigid} & \checkmark  & 0.232  & 0.037    & 59.7            & 82.5            & 91.3            & 97.0            & 99.2              & 99.7     \\\hline
		+Replacement &\checkmark   & 0.168  & 0.032    & 56.5            & 85.7            & 94.4            & 98.4            & 99.7              & 99.8              \\ \hline
        +SPN~\cite{liu2016learning} &\checkmark & 0.162    & 0.027      & 67.5            & 87.9            & 94.7            & 98.5            & 99.7              & 99.9  \\\hline
	    +UNet(Ours)+SPN      & \checkmark &  0.144 & 0.022 & 75.4 & 90.8 & 95.8 & 98.8 & 99.8 & 100.0   \\ \hline
	    +CSPN (Ours)   & \checkmark & 0.136    & 0.021      & 76.2  & 91.2  & 96.2   & 99.0   &99.8     & 100.0   \\ \hline
        +UNet+CSPN (Ours) & \checkmark & \textbf{0.117} & \textbf{0.016} & \textbf{83.2} & \textbf{93.4} & \textbf{97.1} & \textbf{99.2} & \textbf{99.9} & 100.0   \\ \hline
	\end{tabular}
	\egroup
\label{tbl:sota}
\end{table}

\subsection{Comparisons}
\label{subsec:compare}
We compare our methods against various SOTA baselines in terms of the two proposed tasks. (1) Refine the depth map with the corresponding color image. (2) Refine the depth using both the color image and sparse depth samples. For the baseline methods such as SPN~\cite{liu2016learning} and Sparse-to-Dense~\cite{Ma2017SparseToDense}, we use the released code released online from the authors.

\noindent\textbf{NYU v2.} \tabref{tbl:sota} shows the comparison results. Our baseline methods are the depth output from the network of~\cite{Ma2017SparseToDense}, together with the corresponding color image.  
At upper part of \tabref{tbl:sota} we show the results for depth refinement with color only. 
At row ``Bilateral'', we refine the network output from~\cite{Ma2017SparseToDense} using bilateral filtering~\cite{barron2016fast} as a post-processing module with their spatial-color affinity kernel tuned on our validation set. Although the output depths snap to image edges (\figref{fig:example}(c)), the absolute depth accuracy is dropped since the filtering over-smoothed original depths. At  row ``SPN'', we show the results filtered with SPN~\cite{liu2017learning}, using the author provided affinity network. Due to joint training, the depth is improved with the learned affinity, yielding both better depth details and absolute accuracy. Switching SPN to CSPN (row ``CSPN'') yields relative better results.
Finally, at the row ``UNet'', we show the results of just modifying the network with mirror connections as stated in~\secref{subsec:unet}.  The results turn out to be even better than that from SPN and CSPN, demonstrating that by simply adding feature from beginning layers, the depth can be better learned.

At lower part of \tabref{tbl:sota}, we show the results using both color image and sparse depth samples, and all the results preserves the sparse depth value provided. We randomly select 500 depth samples per image from the ground truth depth map. 

\begin{figure}[t]
\centering
\includegraphics[width=0.98\textwidth]{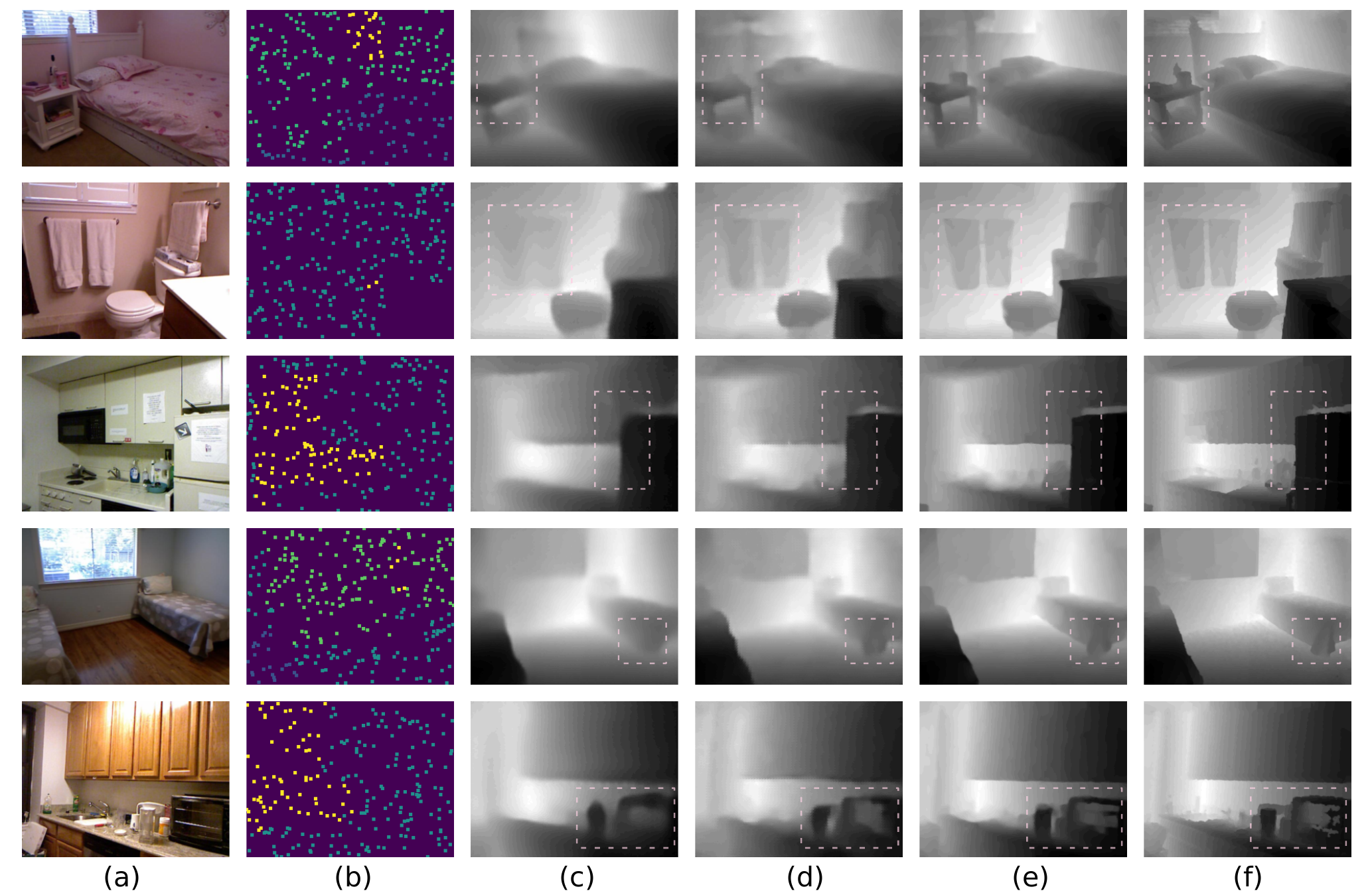}
\caption{Qualitative comparisons on NYU v2 dataset. (a) Input image; (b) Sparse depth samples(500); (c) Ma \etal \cite{Ma2017SparseToDense}; (d) UNet(Ours)+SPN\cite{liu2016learning}; (e) UNet+CSPN(Ours); (f) Ground Truth. Most significantly improved regions are highlighted with yellow dash boxes (best view in color).}
\label{fig:nyudepth}
\end{figure}

For comparison, we consider a baseline method using as-rigid-as-possible (ASAP)~\cite{igarashi2005rigid} warping. Basically the input depth map is warped with the sparse depth samples as control points. At row ``ASAP'', we show its results, which just marginally improves the estimation over the baseline network. For SPN, we also apply the similar replacement operation in \equref{eqn:cspn_sp} for propagation, and the results are shown at row ``SPN'', which outperforms both the results form ASAP and SPN without propagation of SD due to joint training helps fix the error of warping.  At row ``UNet + SPN'', we use our UNet architecture for learning affinity with SPN, which outperforms ``SPN'', while we did not see any improvements compared with that only using UNet. 
Nevertheless, by replacing SPN with our CSPN, as shown in row ``UNet + CSPN'', the results can be further improved by a large margin and performs best in all cases. We think this is mostly because CSPN updates more efficiently than SPN during the training. 
Some visualizations are shown in \figref{fig:nyudepth}. We found the results from CSPN do capture better structure from images (highlighted with dashed bounding boxes) than that from other state-of-the-art strategies.

\begin{figure}[!htpb]
\centering
\includegraphics[width=0.95\textwidth]{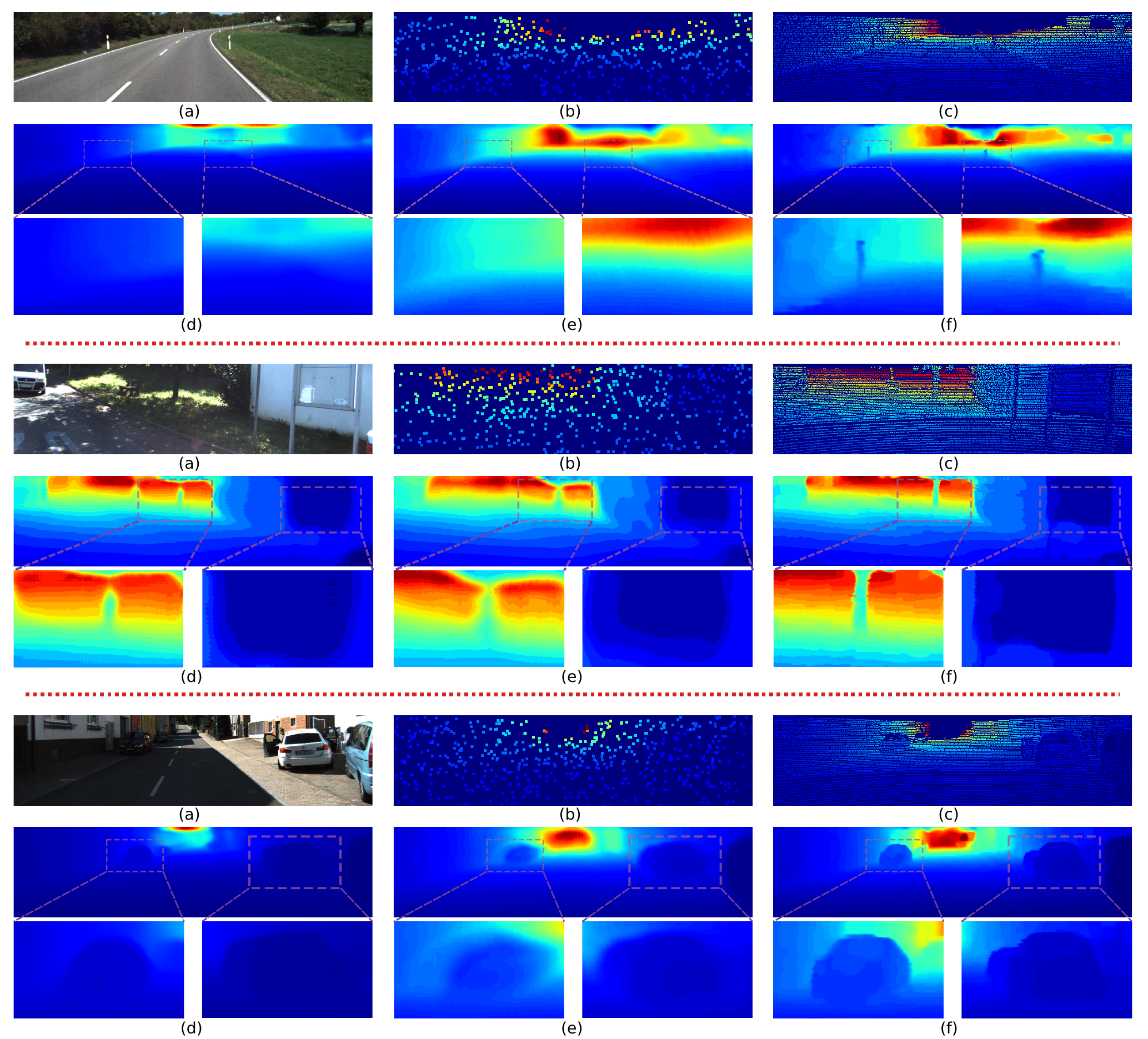}
\caption{Qualitative comparisons on KITTI dataset. (a) Input image; (b) Sparse depth samples(500); (c) Ground Truth; (d) Ma \etal \cite{Ma2017SparseToDense}; (e) Ma\cite{Ma2017SparseToDense}+SPN\cite{liu2016learning};(f) UNet+CSPN(Ours). Some details in the red bounding boxes are zoomed in for better visualization (best view in color).}
\label{fig:kitti}
\end{figure}

\noindent\textbf{KITTI.} \tabref{tbl:sota_kitti} shows the depth refinement with both color and sparse depth samples. Ours final model ``UNet + CSPN'' largely outperforms other SOTA strategies, which shows the generalization of the proposed approach. For instance, with a very strict metric $\delta < 1.02$, ours improves the baseline~\cite{Ma2017SparseToDense} from $30\%$ to $70\%$, which is more than $2\times$ better. More importantly, CSPN is running very efficiently, thus can be applied to real applications.
Some visualization results are shown at the bottom in \figref{fig:kitti}. Compared to the network outputs from~\cite{Ma2017SparseToDense} and SPN refinement, CSPN sees much more details and thin structures such as poles near the road (first image (f)), and trunk on the grass (second image (f)). For the third image, we highlight a car under shadow at left, whose depth is difficult to learn. We can see SPN fails to refine such a case in (e) due to globally vast lighting variations, while CSPN learns local contrast and successfully recover the silhouette of the car. Finally, we also submit our results to the new KITTI depth completion challenge~\footnote{\url{http://www.cvlibs.net/datasets/kitti/eval_depth.php?benchmark=depth_completion}} and show that our results is better than previous SOTA method~\cite{uhrig2017sparsity}. 

\begin{table}[!htpb]
	\centering
	\caption{Comparison results on KITTI dataset~\cite{geiger2012we} 
	}
	\label{tbl:sota_kitti}
	\fontsize{7.5}{7.5}\selectfont
	\bgroup
	\def\arraystretch{1.3}
	\setlength{\tabcolsep}{4.5pt} 
    \begin{tabular}{lccccccccc}
	\hline
		\multicolumn{1}{l}{\multirow{2}{*}{Method}}  & \multirow{2}{*}{Preserve ``SD''} & \multicolumn{2}{c}{Lower the better}   & \multicolumn{6}{c}{Higher the better} \\ \cline{3-10}
		\multicolumn{1}{l}{} & & RMSE  & \multicolumn{1}{c|}{REL} & $\delta_{1.02}$ & $\delta_{1.05}$ & $\delta_{1.10}$ & $\delta_{1.25}$ & $\delta_{1.25^2}$ & $\delta_{1.25^3}$ \\ \hline
	~~Ma \etal ~\cite{Ma2017SparseToDense}                    &                                 & 3.378                                 & 0.073                    & 30.0            & 65.8            & 85.2            & 93.5            & 97.6              & 98.9              \\ \hline
	+SPN~\cite{liu2016learning}                   & \checkmark                                & 3.243                                 & 0.063                    & 37.6            & 74.8            & 86.0            & 94.3            & 97.8              & 99.1              \\ \hline
	+CSPN(Ours)            & \checkmark                                & 3.029                                 & 0.049                    & 66.6            & 83.9            & 90.7            & 95.5            & 98.0              & 99.0              \\ \hline
	+UNet(Ours)            &                              & 3.049                                 & 0.051                    & 62.6            & 83.2            & 90.2            & 95.3            & 97.9              & 99.0              \\ \hline
	+UNet(Ours)+SPN        & \checkmark                               & 3.248                                 & 0.059                    & 52.1            & 79.0            & 87.9            & 94.4            & 97.7              & 98.9              \\ \hline
	+UNet+CSPN(Ours)       & \checkmark                                & \textbf{2.977}                        & \textbf{0.044}           & \textbf{70.2}   & \textbf{85.7}   & \textbf{91.4}   & \textbf{95.7}   & \textbf{98.0}     & \textbf{99.1}     \\ \hline
    \end{tabular}
	\egroup
\label{tbl:sota_kitti}
\end{table}

\section{Conclusion}
In this paper, we propose convolutional spatial propagation network (CSPN), which can be jointly learned with any type of CNN.  It can be regarded as a linear diffusion process with guarantee to converge. Comparing with previous spatial propagation network~\cite{liu2017learning} which learns the affinity, CSPN is not only more efficient (2-5$\times$ faster), but also more accurate (over $30\%$ improvement) in terms of depth refinement. We also extend CSPN by embedding sparse depth samples into the propagation process, which provides superior improvement over other SOTA methods~\cite{Ma2017SparseToDense}. Since our framework is general, in the future, we plan to apply it to other tasks such as image segmentation and enhancement.

\clearpage

\bibliographystyle{splncs}
\bibliography{egbib}
\end{document}